\documentclass{article}
\usepackage[utf8]{inputenc}

\usepackage{amsmath}
\usepackage{amssymb}
\usepackage{tcolorbox}
\begin{document}
\title{Markov Chain Concentration with an Application in Reinforcement Learning}

\author{Debangshu Banerjee }
\date{February 2021}
\maketitle

\begin{abstract}
    Given $X_1,\cdot ,X_N$ random variables whose joint distribution is given as $\mu$ we will use the Martingale Method to show any Lipshitz Function $f$ over these random variables is subgaussian. The Variance parameter however can have a simple expression under certain conditions. For example under the assumption that the random variables follow a Markov Chain and that the function is Lipschitz under a Weighted Hamming Metric. We shall conclude with certain well known techniques from concentration of suprema of random processes with applications in Reinforcement Learning
\end{abstract}

\section{Introduction}
In the class we have been introduced to the concentration of measure for functions of independent random variables. A natural extension of this would be to consider the concentration of measure phenomenon for dependent random variables. In this report we shall investigate this. We shall see how Markov Chain concentration form an example in this setting. We shall highlight the following result which has already been highlighted in a form for independent random variables within the class lectures

\begin{center}
    \textit{Lipschitz Functions defined on Markov Chains are Sub-Gaussian under certain Ergodic Properties of the Markov Chain}
\end{center}

The way this report is organized is as follows:
\begin{itemize}
    \item Theoretical Platform and Definitions
    \item Martingale Method for Concentration of Measure
    \item Wasserstein Matrix
    \item General Result and Extensions
    
\end{itemize}

We shall also present a hypothesis of how concentration of Markov Chains may be utilized in the field of Reinforcement Learning. These results are still in their infancy but we shall nevertheless include them here with hopes of future work.

\section{Theoretical Platform and Definitions}
We shall assume the following : Let $T$ be a countable index set, $(\Omega_t, \mathcal{B}_t)$ be a measurable space where 
$\Omega_t$ is a Polish Space for each $t \in T$ and $d_t$ be a metric function for each $\Omega_t$. 
\par We shall define the product space $\Omega = \otimes_t \Omega_t$, the product sigma algebra $\mathcal{B} = \otimes_t\mathcal{B}_t$ and define a borel probability measure $\mu$ on $\Omega$ (Note that that is not a product measure). We define the metric $d$ on the product space as $d = \sum_t d_t$ 
\\

\textbf{Local Oscillation of $f$} Let $f:\Omega\rightarrow \mathrm{R}$ be a measurable function. We will define the \textit{Local oscillation of $f$} at $i \in T$ as $\Delta_i(f) = \sup_{\substack{x,y \\ x^{-i} = y^{-i}}}\frac{f(x) - f(y)}{d_i(x_i,y_i)}$
\\ That is we keep all except the $i^{th}$ coordinate fixed and vary the $i$ coordinate in forming the ratio.
\\

\textbf{Note}: This definition is siimilar to the bounded difference assumption that was introduced in the class.\\
\textbf{Note}: Under this definition $f(x)-f(y) \leq \sum_{i \in T} \Delta_i(f)d_i(x_i,y_i)$.\\
\textbf{Note}: Under the assumtption that $f$ is Lipshitz under the weighted Hamming Distance $|f(x) - f(y)| \le \sum_{i \in T} c_i \mathrm{1}_{x_i \neq y_i}$, we have $\Delta_i(f) \le c_i$. \\
\\

\textbf{Markov Kernel} $K: \Omega \times \mathcal{B}\rightarrow [0,1]$ is a Markov Kernel if $K$ is a Borel Probability measure for each $x \in \Omega$.\\

\textbf{Note}: If $f$ is any Bounded Measurable function we have $Kf(x) = \int f(y)K(x,dy)$ a measurable function for almost every $x$ in $\Omega$ \\
\textbf{Note}: If $\mu$ is any borel probability measure we can define $\mu K(\mathcal{A}) = \int K(x,\mathcal{A})d\mu(x)$ a borel probability measure $\forall \mathcal{A} \in \mathcal{B}$\\

The following result has been done in class and is presented here without proof:\\
\textbf{Azuma-Hoeffding}: Let $\{\mathcal{F}_k\}_{k\le n}$ be any filtration and $M_k$ be a martingale difference sequence such that $A_k \le M_k \le B_k$ a.s., then $\sum_{k=1}^N M_k$ is subgaussian with variance paramter $1/4 \sum_{k=1}^N||B_k - A_k||_{\infty}^2$

\section{Martingale Method for Concentration of Measure}
Here we see how to use the Martingale Method introduced in class to get the concentration of measure phenomenon. The proof will will take the following course.
\begin{enumerate}
    \item Construct the Martingales\\
    $\mathbf{E}[f(X_1,\cdots,X_n| X_1,\cdots,X_i] - \mathbf{E}[f(X_1,\cdots,X_n| X_1,\cdots,X_{i-1}]$
    
    \item Bound each term above as $f(x_1,\cdots,x_i,X_{i+1},\cdots,X_N) - f(x_1,\cdots,X_i,\cdots,X_N)$ under the assumption of existence of an \textbf{Wasserstein Matrix}
    
    \item Finally use Azuma-Hoeffding to get bounds in term of a matrix norm of this Wasserstein Matrix
    
    \item Motivate how Wassertein Matrices can be represented as Upper Triangular Matrices and get simple interpretations under stronger assumptions.
\end{enumerate}

Thus let us begin
\subsection{Construct the Martingale Differences $M_k$}
Given any borel probability $\mu$ measure on $\Omega$. (Recall $\Omega$ is the product space) let us define $\{X_i\}_{i \in T}$ random variables defined on each $\Omega_i$ with joint distribution $\mu$. Let us define the natural filtration $\{\mathcal{F}_k\}_{i \in T}$. For each $i \in T$ we definte the Markov Kernel 
$$K_i(x,dy) = \delta_{x^{[i-1]}}(dy^{[i-1]})\otimes\mu^{[i,n]}(dy^{[i,n]}|x^{[i-1]})$$. 
That is $K_i$ is nothing but the conditional measure of $\mu$ given $\mu_{[i-1]}$. Note that $[n] = \{1,\cdots,n\}$.\\

Under this definition of the Markov Kernel we have for any bounded measurable function $f : \Omega \rightarrow \mathrm{R}$,
$$K_if(x) = \mathbf{E}_\mu[f(X)|X^{[i-1]} = x^{[i-1]}]$$\\

Now let us denote the Martingale Difference under the natural filtration 
\begin{multline}
    M_i = K_{i+1}f(x) - K_i(f) \\ = \mathbf{E}_\mu[f(X)|X^{[i]} = x^{[i]}] - \mathbf{E}_\mu[f(X)|X^{[i-1]} = x^{[i-1]}] 
\end{multline}
 Therefore Note that $f(X_1,\cdots,X_N) - \mathbf{E}[f(X_1,\cdots,X_N)] = \sum_{i=1}^N M_i$
 
\subsection{Bounding the Martingale differences}
Using the tower property of conditional expectations 
$$M_i = \mathbf{E}_\mu[f(X)|X^{[i]} = x^{[i]}] - \mathbf{E}_\mu[f(X)|X^{[i-1]} = x^{[i-1]}]$$
$$= \mathbf{E}_\mu[f(X)|X^{[i]} = x^{[i]}] - \mathbf{E}_\mu[\mathbf{E}_\mu[f(X)|X^{[i-1]} = x^{[i-1]}, X_i]|X^{[i-1]} = x^{[i-1]}]$$
\begin{multline}
 =   \int_{\Omega_{[i,n]}}\big(\int_{\Omega_{(i,n]}}f(x_{[i-1]}x_iy_{(i,n]})\mu^{(i,n]}(dy^{(i,n]}|x^{[i]}) -  \int_{\Omega_{(i,n]}}f(x_{[i-1]}y_iy_{(i,n]})\mu^{(i,n]}(dy^{(i,n]}|x^{[i]},y_i)\big)\\\mu^{[i,n]}(dy^{[i,n]}|x^{[i-1]})
\end{multline}

\begin{multline}
 =   \int_{\Omega_{[i,n]}}\big(K_{i+1}f(x_{[i-1]}x_iy_{(i,n]})-  K_{i+1}f(x_{[i-1]}y_iy_{(i,n]})\big)\mu^{[i,n]}(dy^{[i,n]}|x^{[i-1]})
\end{multline}

This implies that each $M_i$ can be bounded by 
$$A_i = \int_{X_{[i,n]}}\inf_{x_i \in X_i}\big(K_{i+1}f(x_{[i-1]}x_iy_{(i,n]})-  K_{i+1}f(x_{[i-1]}y_iy_{(i,n]})\big)\mu^{[i,n]}(dy^{[i,n]}|x^{[i-1]})$$
$$B_i = \int_{X_{[i,n]}}\sup_{x_i \in X_i}\big(K_{i+1}f(x_{[i-1]}x_iy_{(i,n]})-  K_{i+1}f(x_{[i-1]}y_iy_{(i,n]})\big)\mu^{[i,n]}(dy^{[i,n]}|x^{[i-1]})$$

Now using the bounded differnces definition of $\Delta_i(K_{i+1}f)$, we have 
$$||B_i - A_i||_\infty < ||d_i||\Delta_i(K_{i+1}f)$$ where $||d_i|| = \sup_{x_i,z_i}d_i(x_i,z_i)$\\

So now we are in the vicinity of using Azuma-Hoeffding. The only remaining step remains is how to bound $\Delta_i(K_{i+1}f)$ using some known function of $\Delta_i(f)$.

\section{Wasserstein Matrices}
From the theory of general contractive Markov Kernels $K$ with Dobrushin Coefficient $\theta$  defined as:
$$\sup_{x,y}||K(x,.) - K(y,.)||_{TV} < \theta$$,
under the Weighted Hamming Distance $\forall f \in Lip(X,d)$ 
$$\Delta_i(Kf) < \frac{\theta}{\alpha_i}\sum_{j\in T}\alpha_j\Delta_j(f)$$\\
We refer to \cite{kontorovich2017concentration} for a proof of the above.\\

As displayed above it is suggested that all $\Delta_j(f)$ $\forall j \in T$ influence $\Delta_i(Kf)$. With such a relation as a motivation a \textbf{Wasserstein Matrix} is defined:\\

\textbf{Wasserstein Matrix}A Markov Kernel $K$ is said to have a \textit{Wasserstrien Matrix} $\mathbf{V} = (V_{ij})_{i,j \in T}$ $V_{ij} >0 $ if $\forall f \in Lip(X,d)$ 
$$\Delta_i(Kf) < \sum_{j\in T}V_{ij}\Delta_j(f) \forall i\in T$$
or in vector form
$$\mathbf{\Delta(Kf)} < \mathbf{V\Delta(f)}$$

\section{General Result and Extensions}
\subsection{SubGaussianity of Markov Kernels under assumption of existence of Wassertein Matrix}

Let us make the following assumption
\begin{center}
    For each Markov Kernel $K_i$, $i \in T$ defined as before assume there exists Wasserstein Matrices $\mathbf{V^i} = (V^i_{lm})_{l,m \in T}$ s.t for each $i \in  T$ $$\mathbf{\Delta(K_if)} < \mathbf{V^i\Delta(f)}$$ for each $f\in Lip(X,d)$
\end{center}
What this assumption is essentially saying is a rather strong statement. It states that we assume that for each markov kernel $K_i$ as defined before, there exists a Wasserstein Matrix $\mathbf{V^i}$. We shall soon see how we can back up our assumption by explicit construction of Wasserstein Matrices for general probability measures $\mu$, and later give simple expressions of Wasserstein Matrices under further assumptions about the structure of $\mu$ and the distance metric $d$.\\
For the time being let us see how this assumption immediately gives us a SubGaussianity result\\

As before, from Azuma-Hoeffidfing, $f(X_1,....X_N) - \mathbf{E}f(X_1,...X_N)$ is subgaussian with variance paramter $\sum_{k=1}^N||B_k - A_k||^2_\infty$
where 
$$\sum_{k=1}^N||B_k - A_k||^2_\infty <= \sum_{k=1}^N||d_k||^2\Delta_k(K_{k+1}f)^2$$
$$=\sum_{k=1}^N(\sum_{j \in T}||d_k||V^{k+1}_{kj}\Delta_j(f))^2$$
$$= \sum_{k=1}^N(\sum_{j \in T}\Gamma_{kj}\Delta_j(f))^2$$
$$= \sum_{k=1}^N\mathbf{\Gamma\Delta(f)}_k^2 = ||\mathbf{\Gamma\Delta(f)}||^2_{l_2(T)}$$
    where we define $\Gamma_{ij} = ||d_i||V^{i+1}_{ij}$
\\

The remaining of the document is sectioned as follows:
\begin{itemize}
    \item Construction of Wassersterin Matrices Using Couplings
    \item Wassersterin Matrices under the discrete Metric and Goldstein Coupling
    \item Contractive Markov Chains
    \item Uniformly Ergodic Markov Chains
    
\end{itemize}

\subsection{Construction of Wassersterin Matrices Using Couplings}
This section illustrates the most general form of a Wasserstein Matrix.\\

Let $\mathbf{P}^{[i]}_{xy}$ be any coupling of conditional probabilities $\mu^{(i,n]}(.|x^{[i]})$ and $\mu^{(i,n]}(.|y^{[i]})$ where $x$ and $y$ only differ in the $i$ coordinate.
Then recall that we are trying to relate $\Delta_i(K_{i+1}f)$, we see
\begin{multline}
K_{i+1}f(x) - K_{i+1}f(y) = \\
\int_{X_{(i,n]}}\int_{X_{(i,n]}} (f(x^{[i]},u^{(i,n]}) - f(y^{[i]},v^{(i,n]})\mu^{(i,n]}(du^{(i,n]}|x^{[i]})\mu^{(i,n]}(dv^{(i,n]}|y^{[i]})\\
= \int_{X_{(i,n]}\times X_{(i,n]}}\mathbf{P}^{[i]}_{xy}(du^{(i,n]}, dv^{(i,n]})(f(x^{[i]},u^{(i,n]}) - f(y^{[i]},v^{(i,n]})   
\end{multline}

Further note that by our construction of the Markov Kernels $K_i$, we have
$\Delta_i(K_jf) = 0$ for $ i>j$
Thus we can express the above as 
\begin{multline}
\leq \Delta_i(f)d_i(x_i,y_i) + \\ \sum_{j>i}\Delta_j(f)\int_{X_{(i,n]}\times X_{(i,n]}}\mathbf{P}^{[i]}_{xy}(du^{(i,n]}, dv^{(i,n]}) d_j(u_j,v_j)
\end{multline}
or,
$$\frac{K_{i+1}f(x) - K_{i+1}f(y)}{d_i(x_i,y_i)}\leq \Delta_i(f) +  \sum_{j>i}\frac{\int\mathbf{P}^{[i]}_{xy}d_j}{d_i(x_i,y_i)}\Delta_j(f)$$
and therefore,
$$\Delta_i(K_{i+1}f)\leq \Delta_i(f) +  \sum_{j>i}\sup_{\substack{x,y \\ x^{-i} = y^{-i}}}\frac{\int\mathbf{P}^{[i]}_{xy}d_j}{d_i(x_i,y_i)}\Delta_j(f)$$
Thus from our definition of Wasserstien Matrices
\[
V^{i+1}_{ij} = 
\begin{cases}
0 &\quad\text{if }i>j \\
1 &\quad\text{if }i = j \\
\sup_{\substack{x,y \\ x^{-i} = y^{-i}}}\frac{\int\mathbf{P}^{[i]}_{xy}d_j}{d_i(x_i,y_i)} &\quad\text{if }i < j
\end{cases}
\]

\subsection{Wassersterin Matrices under the discrete Metric and Goldstein Coupling}
Let us introduce a further assumption, that we are working with a discrete metric.\\

Under the discrete metric,
$$\Gamma_{ij} = \sup_{\substack{x,y \\ x^{-i} = y^{-i}}} \mathbf{P}^{[i]}_{xy}[Y^{(0)}_j \neq Y^{(1)}_j]$$
where $(Y^{(0)}, Y^{(1)}) = ((Y^{(0)}_{i+1},\cdots, Y^{(0)}_n), (Y^{(1)}_{i+1},\cdots, Y^{(1)}_n)$ is a random variable taking value in $X_{(i,n]}\times X_{(i,n]}$ with $Y^{(0)}$ distributed as $\mu^{(i,n]}(.|x^{[i]})$ and $Y^{(1)}$ distributed as $\mu^{(i,n]}(.|y^{[i]})$
$$\mathbf{P}^{[i]}_{xy}[Y^{(0)}_j \neq Y^{(1)}_j] \leq 
\mathbf{P}^{[i]}_{xy}[(Y^{(0)}_j,\cdots,Y^{(0)}_n) \neq (Y^{(1)}_j,\cdots,Y^{(1)}_n)]$$

\textbf{Goldstein Coupling}\cite{marton2004measure}\\
There exists a coupling $\mathbf{P}^{[i]}_{xy}$ known as the Goldstein Maximal Coupling s.t. $\mathbf{P}^{[i]}_{xy}[(Y^{(0)}_j,\cdots,Y^{(0)}_n) \neq (Y^{(1)}_j,\cdots,Y^{(1)}_n)] = ||\mu^{[j,n]}(.|x^{[i]}) - \mu^{[j,n]}(.|y^{[i]})||_{TV}$\\

Thus using the Goldstein Coupling
\[
\Gamma_{ij} = 
\begin{cases}
0 &\quad\text{if }i>j \\
1 &\quad\text{if }i = j \\
\sup_{\substack{x,y \\ x^{-i} = y^{-i}}}||\mu^{[j,n]}(.|x^{[i]}) - \mu^{[j,n]}(.|y^{[i]})||_{TV} &\quad\text{if }i < j
\end{cases}
\]

\subsection{Contractive Markov Chains}
Let us revisit the example of Contractive Markov Chain\\

Here we have a directed Markov Model, where 
$\mu = \mu_0K_1....K_n$ where $K_i$ is a Markov Kernel from $X_i$ to $X_{i+1}$
under the Doeblin Contraction Coefficient
$$\sup_{x,y}||K_i(x,.) - K_i(y,.)||_{TV} < \theta_i$$
Under this assumption it can be shown using a recursion argument that,
$$||\mu^{[j,n]}(.|x^{[i-1]},x_i) - \mu^{[j,n]}(.|x^{[i]}, y_i)||_{TV} \leq \theta^{j-i}$$ where $\theta = \max \theta_i$.\\
We refer to \cite{kontorovich2012obtaining} or \cite{samson2000concentration} for a full proof.\\
 Thus from the previous sections, we have
 \[
\Gamma =
\begin{pmatrix}
1       &\theta     &\theta^2   &\dots  &\theta^{n-1}\\
0       &1          &\theta     &\dots  &\theta^{n-2}\\
\vdots  &\vdots     &\vdots     &\ddots & \vdots \\
0       &0          &0          &\dots   &1
\end{pmatrix}
\]
Now from our general result, and noting that if $f$ is Lipshitz under the weighted Hamming Distance $|f(x) - f(y)| \le \sum_{i \in T} c_i \mathrm{1}_{x_i \neq y_i}$, we have $\Delta_i(f) \le c_i$.
we have subgaussianity with
$||\Gamma||^2||c||^2 $ where $c = (c_i)_{i\le n}$

\subsection{Uniformly Ergodic Markov Chains}
Be recalling a general result that uniform ergodic Markov Chains have finite mixing times, we have the following variant of the general result.\\

Let us first define what the mixing time for a Markov Chain is:
$$\tau(\epsilon) = \min_t[\{\max_{1<i<N-t}\max_{x,y}||\mathbf{P}(X_{i+t}|X_i=x) - \mathbf{P}(X_{i+t}|X_i=y)||_{TV}\} \le \epsilon]$$
Let $n = N/\tau(\epsilon)$ and let us form $n$ partitions for the Markov Process of size $\tau(\epsilon)$. The main theorem that was presented also holds for such partitions by redefining the $\sigma-algebra$ defined in the proof. Also we note for this partitioned Markov Chain we have $$||\mu^{[j,n]}(.|\hat{x}^{[i-1]},\hat{x}_i) - \mu^{[j,n]}(.|\hat{x}^{[i]}, \hat{x}_i)||_{TV} \leq \theta^{j-i-1} $$. We refer to \cite{paulin2015concentration} for a full proof.\\
This gives the matrix $\Gamma$ as
\[
\Gamma =
\begin{pmatrix}
1       &1     &\epsilon   &\dots  &\epsilon^{n-2}\\
0       &1          &1     &\dots  &\epsilon^{n-3}\\
\vdots  &\vdots     &\vdots     &\ddots & \vdots \\
0       &0          &0          &\dots   &1
\end{pmatrix}
\]

Therefore for any Lipschitz function w.r.t the weighted discrete metric would be subgaussian with Variance Parameter $\tau(\epsilon)||c||^2||\Gamma||^2$

\section{Reinforcement Learning: A Hypothesis}
We conclude this section by perhaps an interesting phenomenon of how such Markov Chain concentrates can be used in the field of reinforcement Learning.\\

Recall that for any policy $\pi$ the evolution of states $X_i$ and rewards $r_i$ follow a Markov Process.

In reinforcement Learning we are interested in the cumulative reward over a fixed time which is a function of the trajectory $X_1, \cdots , X_N$. 
\begin{tcolorbox}
Let us have a Lipschitz function $V_\pi(X_1,\cdots,X_N) \rightarrow \mathbf{R}$.  Let us assume the underlying Markov Process $X_1, \cdots , X_N$ is ergodic under the policy $\pi$ then we have the rather nice result that the $V_\pi(X_1,\cdots,X_N)$ is subgaussian. Let us denote the variance proxy by $\sigma_\pi^2$. 
\end{tcolorbox}
In reinforcement Learning we are interested in the $\mathbf{E}sup_{\pi \in \Pi}V_\pi$ where $\Pi$ is a class of policies.

Under a very trivial case if we assume that $\Pi$ contains only finite number of policies then under our hypotheses of $V_\pi$ is subgaussian with variance parameter $\sigma_\pi^2$ for any policy $\pi$.\\

Then using a \textbf{Maximal Inequality} for subgassians
$$\mathbf{E}sup_{\pi \in \Pi}V_\pi \le \sqrt{2\sigma^2\log |\Pi|}$$
where $\sigma = \max \sigma_\pi $

This bound is sharp under assumption of independence. We refer to an exercise problem in \cite{van2014probability} for a more general result.
\\

Under the condition $\Pi$ is not finite we have to introduce a sense of continuity for the map $\pi\rightarrow V_\pi$.
The simplest such sense would be to consider a \textbf{Lipschitz Process}\\

Let $\Pi$ be equipped with a metric $d$ and assume $(V_\pi)_{\pi \in \Pi}$ is a \textbf{Lipschitz Process} i.e.
$$|V_\pi - V_{\pi'}| < C d(\pi, \pi') a.s.$$, where $C$ is a random variable
Then 
\begin{tcolorbox}
Under the assumption of $(V_\pi)_{\pi \in \Pi}$ is a \textbf{Lipschitz Process} on $(\Pi, d)$ with $V_\pi$ subgaussian with variance proxy $\sigma^2$ for every $\pi \in \Pi$, then  
$$\mathbf{E}sup_{\pi \in \Pi}V_\pi \le \inf_\epsilon \{\epsilon\mathbf{E}[C] + \sqrt{2\sigma^2\log N(\Pi,d,\epsilon)}\}$$
where $N(\Pi,d,\epsilon)$ is the covering number for $\epsilon$-nets.
\end{tcolorbox}

This result is not sharp.

The sharpest known inequality gives a rather interesting result\\

\textbf{Dudley's Inequality and SubGaussian Processes}
Let $(V_\pi)_{\pi \in \Pi}$ on $(\Pi, d)$ be a subgaussian process:
$$\mathbf{E}[\exp{\lambda(V_\pi - V_{\pi'})}] \leq \exp{\lambda^2d(\pi,\pi')^2/2}$$

That we indeed have subgaussian processes for Markov Chains can perhaps be noted from the following observation
\begin{tcolorbox}
Note that under our hypothesis of $V_\pi$ subgaussian $\mathbf{E}[\exp{\lambda V_\pi}] < \exp{\lambda^2\tau_\pi(\epsilon)}$. Thus it seems probable that our natural metric induced on the class of policies could be mixing times under various policies.
\end{tcolorbox}

\begin{tcolorbox}[sharp corners, colback=green!30, colframe=green!80!blue, title=Dudley's Inequality]
For separable subgaussian processes
$$\mathbf{E}sup_{\pi \in \Pi}V_\pi \le  12 \int_{0}^\infty\sqrt{\log N(\Pi,d,\epsilon)}d\epsilon$$
\end{tcolorbox}

\subsection{Further hypotheticals}

\textbf{Value Functions}\\

For Value Functions we define $$V_\pi(X_1,\cdots,X_N) = r(X_1, \pi(X_1)) + \cdots + r(X_N, \pi(X_N))$$. Here $r(X_i, \pi(X_i))$ is a random variable and denotes the reward incurred under the $i^{th}$ stage at state $X_i$ under policy $\pi$. Let us assume that rewards at any stage $i$ is nonnegative and bounded by $r_i$. That is\\
$0 \le r(X_i, \pi(X_i)) \le r_i$ $\forall X_i \in \Omega_i$ and for any $\pi$. \\

Then $$V_\pi(x_1,\cdots,x_N) - V_\pi(y_1,\cdots,y_N) = \sum_{i=1}^N(r(x_i, \pi(x_i)) - r(y_i, \pi(y_i)))$$
$$ \leq \sum_{i=1}^Nr_i\mathrm{1}_{\{x_i \neq y_i\}}$$ for any policy $\pi$
This gives that the Value function is Lipshitz under the weighted Hamming Metric and our previous hypothesis would apply verbatim.\\

\textbf{Finite State-Action Space}\\

Let State Space be of cardinality $S$, action-space be of cardinality $A$ and the time horizon be $H$. This would suggest that the number of policies available would be no more than $SA$. Since this is a finite set, we can apply the maximal inequality for subgaussians and conclude
\begin{tcolorbox}[sharp corners, colback=green!30, colframe=green!80!blue, title=Finite State Space Hypothesis]
Under the assumption that rewards are bounded between $[0,1]$ 
$$\mathbf{E}sup_{\pi \in \Pi}V_\pi \le  \sqrt{H\tau_{mix}\log SA}$$
\end{tcolorbox}
\textbf{Proof Sketch}: Note that $V_\pi$ is subgaussian with variance paramater $H \tau_\pi$. Let
where $\tau_{mix}$ be the maximum mixing time of all such policies. Then we use Maximal Inequality formula.

A trivial upper bound can be established also where instead of finding the maximum of $\tau_\pi$ we just perform an union bound of $SA\tau_\pi$\\
This gives the following bound
\begin{tcolorbox}[sharp corners, colback=green!30, colframe=green!80!blue, title=Finite State Space Hypothesis]
Under the assumption that rewards are bounded between $[0,1]$ 
$$\mathbf{E}sup_{\pi \in \Pi}V_\pi \le  \Tilde{\mathcal{O}}(\sqrt{HSA\tau_\pi})$$ for any policy $\pi$
where $\Tilde{\mathcal{O}}$ supreses the $\log$ terms
\end{tcolorbox}
One can see \cite{ortner2020regret} where an algorithm is given which suffers a similar regret.

\section{Acknowledgements}
I would like to express my thanks to Professor Aditya Gopalan and Prof Himangshu Tyagi. The references are included herein  \cite{levin2017markov}, \cite{raginsky2012concentration}

\medskip

\bibliographystyle{unsrt}
\bibliography{bibliography}

\end{document}